\documentclass[conference,]{IEEEtran}
\IEEEoverridecommandlockouts
\usepackage{cite}
\usepackage{amsmath,amssymb,amsfonts}
\usepackage{algorithmic}
\usepackage{graphicx}
\usepackage{textcomp}
\usepackage{xcolor}
\usepackage{dblfloatfix}
\usepackage{flushend}

\usepackage{multirow}
\usepackage{tablefootnote}
\usepackage[caption=false]{subfig} 

\newcommand{\figref}[1]{Fig.~\ref{#1}} 

\def\BibTeX{{\rm B\kern-.05em{\sc i\kern-.025em b}\kern-.08em
    T\kern-.1667em\lower.7ex\hbox{E}\kern-.125emX}}
\begin{document}

\title{Intent Detection at Scale: \\Tuning a Generic Model using Relevant Intents}

\author{\IEEEauthorblockN{Nichal Narotamo}
\IEEEauthorblockA{\textit{Zendesk} \\
\small nichal.narotamo@zendesk.com}
\and
\IEEEauthorblockN{David Aparício}
\IEEEauthorblockA{\textit{Zendesk} \\
\small david.aparicio@zendesk.com}
\and
\IEEEauthorblockN{Tiago Mesquita}
\IEEEauthorblockA{\textit{Zendesk} \\
\small tiago.mesquita@zendesk.com}
\and
\IEEEauthorblockN{Mariana Almeida}
\IEEEauthorblockA{\textit{Zendesk} \\
\small mariana.almeida@zendesk.com}
}

\maketitle

\begin{abstract}
Accurately predicting the intent of customer support requests is vital for efficient support systems, enabling agents to quickly understand messages and prioritize responses accordingly.  While different approaches exist for intent detection, maintaining separate client-specific or industry-specific models can be costly and impractical as the client base expands.

This work proposes a system to scale intent predictions to various clients effectively, by combining a single generic model with a per-client list of relevant intents. Our approach minimizes training and maintenance costs while providing a personalized experience for clients, allowing for seamless adaptation to changes in their relevant intents. Furthermore, we propose a strategy for using the clients relevant intents as model features that proves to be resilient to changes in the relevant intents of clients -- a common occurrence in production environments.
The final system exhibits significantly superior performance compared to industry-specific models, showcasing its flexibility and ability to cater to diverse client needs. 
\end{abstract}


\renewcommand{\arraystretch}{1.2}

\section{Introduction}

\begin{figure*}[t]
    \centering

    \begin{minipage}[b]{.3\textwidth}\centering
    \subfloat[Generic model]{
        \includegraphics[width=.95\textwidth]{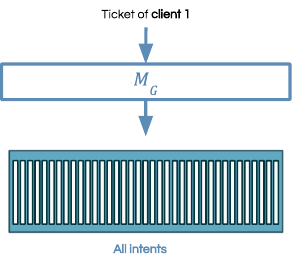} 
        \label{fig:approaches-a}
    }
    \end{minipage}%
    \begin{minipage}[b]{.36\textwidth}\centering
    \subfloat[Industry models]{
        \includegraphics[width=.95\textwidth]{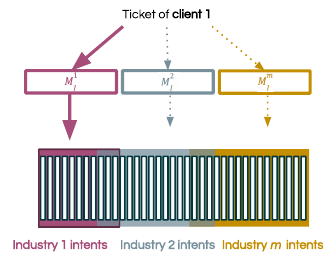}
        \label{fig:approaches-c}
    }
    \end{minipage}%
    \begin{minipage}[b]{.36\textwidth}\centering
    \subfloat[\textbf{Ours}: generic model w/ relevant intents]{
        \includegraphics[width=.95\textwidth]{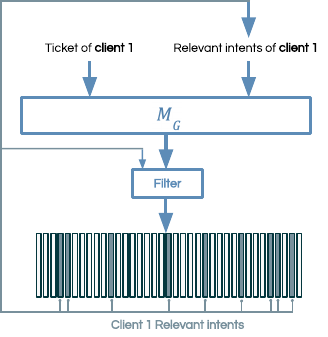} 
        \label{fig:approaches-d}
    }
    \end{minipage}%
    \caption{Approaches for intent detection serving various clients. \protect\subref{fig:approaches-a} A generic model that serves all clients. \protect\subref{fig:approaches-c} Industry-specific models process tickets of clients assigned to the respective industry. \protect\subref{fig:approaches-d} Our approach leverages a single generic model by integrating the clients' relevant intents, enhancing performance through additional input to the model and ensuring valid outputs by employing a client-tailored filter.}
    \label{fig:approaches}
\end{figure*}


Automatically detecting the intent of customer support requests (i.e., \emph{tickets}) is a fundamental aspect of an efficient intelligent support system. When the intent of incoming tickets is known, customer support agents can quickly grasp the tickets content and react accordingly, enhancing their efficiency. 
For instance, intent detection enables the creation of priority queues, helping agents identify the most relevant tickets. 
Furthermore, intent detection 
facilitates the mapping of intents to 
\mbox{(semi-)automated} replies, such as macros, further streamlining the support process~\cite{mesquita2022dense}. Finally, when connected with analytic systems, intent information provides a powerful understanding of the customer support activity and requests.

This paper focuses on the challenge of achieving effective intent detection for customer support at scale. In the context of this challenge, the intent detection system must handle tickets from end-users belonging to multiple clients and industries. Each industry, such as financial institutions, software houses, or e-commerce platforms, typically exhibits a distinct set of relevant intents that need to be accurately classified.


Intent detection is commonly performed using supervised machine learning models for classification.  
The model takes the ticket content as input, which, in the case of an email, can include the email's subject, body, and additional features referring to user or client information. The output corresponds to one or multiple intents from a predefined set. 


To perform intent detection at scale, one possible approach is to have a single model supporting all clients and all their possible intents, which we call a \emph{generic model}.  Despite being cost-effective and easy to maintain, this solution can output out-of-domain intents, causing agents and clients to lose confidence in the system. 
Industry-specific models 
serve clients from similar industries. 
Although this approach offers 
flexibility, it 
poses challenges as the number of industries increases and assigning clients to specific industries becomes progressively more complex. Additionally, clients that fall between multiple industries 
complicate the assignment process and cannot be properly covered by this solution.

Finally, in real-world scenarios, the list of relevant intents for a client evolves over time. 
Similarly, some intents may stop being relevant for a business. This introduces the challenge of adapting the intent detection system to accommodate such changes. It is crucial for the system to remain robust, even if new intents emerge without prior communication before model retraining, to ensure that its performance does not experience a significant decline.


Our proposed system, shown in \figref{fig:approaches} \subref{fig:approaches-d},
features 
a single generic model that considers both the ticket content and a list of the client's most relevant intents. 
The list of relevant intents can be derived from the client's historical data or predefined by the client themselves, and serves as a more detailed representation of the client's industry and specific requirements. This approach also leverages the relevant intents as valuable client information, enhancing the accuracy of intent prediction for each ticket. 
Additionally, a filtering module is employed to eliminate irrelevant intents specific to that client in an agile way that is independent from retraining schedules.




The main contribution of the paper, is the development of a scalable intent detection system which: 
\begin{itemize}
    \item Allows clients to have their own personalized set of intents. 
    \item Removes the need to assign clients to industries and to deploy multiple industry-specific models. 
    \item Has superior performance when compared against a generic model by using the list of relevant intents as features (Section~\ref{sec:robustness}). 
    \item Includes a training procedure that is robust to changes in the list of relevant intents allowing for a fast and flexible way of adjusting per-client relevant intents over time. 
\end{itemize}



\section{Method}
\label{sec:method}

\subsection{Problem formulation}


Let us formulate the intent detection task as a classification problem where the goal is to predict one intent $y$, from a predefined list of intents $\mathcal{I}$, given as input a list of features $X$, which can 
contain a representation of the ticket (e.g., the concatenation of the ticket's subject and description) as well as other features providing additional information, such as client or end-user features. More formally,
\begin{equation}
    \hat{y} = M (X) \in \mathcal{I},
\end{equation}

\noindent where $M$ denotes the model used to perform the prediction, which receives $X$ as input and predicts one intent $\hat{y} \in \mathcal{I}$.


In the customer support domain, various approaches can be employed for intent detection (\figref{fig:approaches}). We assume $n$ clients and $m$ industries, with $m << n$. Arguably the simplest approach uses a single generic model $M_{G}$ that handles all incoming tickets without considering their origin (\mbox{\figref{fig:approaches} \subref{fig:approaches-a}}). 
Alternatively,
we can use 
$m$ industry-specific models where $M_{I}^i$ handles tickets from clients within the respective industry $i$, with intents limited to the relevant intents for that industry, i.e., $\mathcal{I}^i \subseteq \mathcal{I}$ (\figref{fig:approaches} \subref{fig:approaches-c}). In all cases, we assume that the models solely utilize ticket related information, 
denoted as $X = t$, as their input features. More formally,

\begin{equation}
    \begin{split}
        \hat{y} & = M_{G} (t) \in \mathcal{I}, \\
        \hat{y} & = m_i(c, \{M_{I}^0 (t), M_{I}^1 (t),\dotsc, M_{I}^m (t)\}) \in \mathcal{I}^{i_c},
    \end{split}
\end{equation}
\noindent where $m_c$ is a function mapping client $c$ to model $M_{C}^c$ and which obtains the predicted intent for $c$; similarly, $m_i$ is a function mapping client $c$ to an industry $i_c$, which then maps $i_c$ to a model $M_{I}^{i_c}$, and finally gets the predicted intent for $c$.

The aforementioned strategies has its own advantages and limitations. The generic model offers simplicity in terms of training, deployment, and maintenance since only one model is used. However, it may produce irrelevant intents for clients, which can affect client perception. Industry-specific models have the advantage of being tailored to specific use-cases, thus reducing the likelihood of producing irrelevant intents (e.g., predicting the ``request product refund" intent for an client associated to the Healthcare industry). However, these models have associated costs, as previously mentioned, and require client-to-industry assignment which can be challenging when clients do not fit into a single industry category. 


\subsection{Proposed solution}\label{sec:proposed_solution}

We propose a generic model that leverages not only the ticket input $t$ from client $c$ but also incorporates a list of relevant intents $I_c$, i.e., $X = (t, I_c)$ (\figref{fig:approaches} \subref{fig:approaches-d}). Our goal is to combine the strengths of a generic model and client-specific models, by using a single generic model while producing meaningful intents tailored to each client. This approach significantly reduces model training time, deployment time, and production costs. The list of client-specific relevant intents $I_c$ can be obtained through client feedback, where clients suggest new intents to be added, or through automated methods by considering the client's history (e.g., past predictions from a production model), or through both. More formally,

\begin{equation}
    \hat{y} = M_{G} (t, I_c) \in \mathcal{I}.
\end{equation}

Our architecture incorporates a \emph{filtering mechanism} applied after inference to ensure valid outputs. When the model predicts an intent that is not within the list of relevant intents, that prediction passes through the filter. When filtering is enabled, we evaluate the model's performance by either (i) considering the prediction as incorrect or (ii) ignoring the top-1 prediction and selecting the first model prediction that matches an intent in the relevant intents, and then assess if it is correct. We call this second approach \textit{``generic with search"}; in Section~\ref{sec:Results} we clarify which option is chosen in each experiment.

\subsection{Model architecture}

\begin{figure}[t]
    \centering
    \includegraphics[width=0.9\linewidth,]{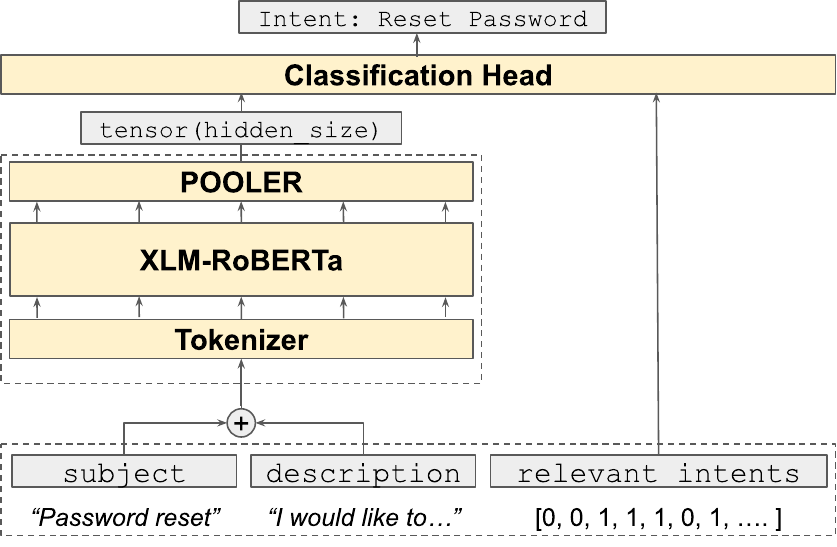} 
    \caption{Proposed solution architecture. The input is the ticket's subject and description, as well as the client's list of relevant intents.}
    \label{fig:model_architecture}
\end{figure}

We employ a transformer encoder architecture, namely XLM-RoBERTa~\cite{liu2019roberta}, to encode the ticket's message. We choose XLM-RoBERTa for its ability to handle multilingual requests, addressing the diverse language requirements in customer support scenarios. In the classification head of our model (\figref{fig:model_architecture}), we enhance it by incorporating the list of relevant intents for the client, providing personalized intent prediction. The prediction is obtained as follows,

\begin{equation}
    \begin{split}
        \hat{y} = {h}\left( \texttt{[CLS]} \oplus I_c \right)
    \end{split},
\end{equation}
\noindent where $\texttt{[CLS]}$
corresponds to the output embedding of the special classification token of XLM-RoBERTa, $h()$ corresponds to the classification head, and $\oplus$ denotes the concatenation operator. In the modified classification head, $h()$, the $\texttt{[CLS]}$ and the list of relevant intents are processed using an Input Mapper, in which the dimensionality of the vectors is reduced. In particular, the $\texttt{[CLS]}$ embedding is multiplied by an identity matrix to keep the same dimension, while the relevant intents array is projected into a 16-dimensional space with embedding dropout. The subsequent module, the Aggregator, combines the embeddings (e.g., through concatenation, sum, or mean). This joint embedding is then fed into a linear projection layer, followed by a configurable number of residual layers.  Each residual layer consists of a linear layer, a dropout layer, and a non-linear layer (e.g., \emph{tanh}). Finally, the resulting embedding is passed through a final classification neural network to provide the intent prediction. More details on the architecture options, such as Aggregator, number of residual layers, etc., are provided in the experimental setup in Section~\ref{sec:hyperparameters}.

\section{Results}
\label{sec:Results}

\subsection{Dataset}
In our experiments we use an in-house customer support dataset comprised of real-world anonimized tickets; 
for data privacy concerns, we are unable to release this dataset. The input for the intent detection models is the concatenation of the ticket's subject 
with its description. The dataset encompasses tickets in nine languages and of 683 different intents related to customer support, such as ``add new item to order", ``delete account", or ``refund request" from various industries. To split the dataset, we employ stratified sampling, allocating 15\% of the data to the validation set while preserving the proportion of examples from each intent class. To account for industry-specific variations, we generate three additional industry-specific datasets. These datasets contain only tickets with intents 
that are valid for the industry. 
The mapping of tickets to intents, of intents to industries, and of clients to industries was done by our in-house experts. 

We create a test set for each dataset by selecting a subset of the validation set that exclusively includes clients manually assigned to that specific industry. This choice was made to ensure that during training, the model learned from tickets that were consistently relevant to the specific intent, while during testing, the focus was on evaluating the model's performance in real production scenarios, where industry-specific models exclusively handle tickets from clients within their respective industries. Additionally, we remove samples from the test set if they have an annotated intent that does not exist in the list of relevant intents for that industry. 

Table~\ref{tab:dataset_details} presents relevant statistics for four datasets (generic dataset and three datasets corresponding to three industries). 

\begin{table}[t]
\centering
\caption{Number of tickets, intents, and clients in the generic dataset and each industry subset.}
\label{tab:dataset_details}
\resizebox{\columnwidth}{!}{%
\begin{tabular}{rl|rrrr}
 \multicolumn{1}{l}{}& \multicolumn{1}{l}{\bf Split}& \multicolumn{1}{|l}{\bf Generic} & \multicolumn{1}{l}{\bf Software} & \multicolumn{1}{l}{\bf E-commerce} & \multicolumn{1}{l}{\bf Finance} \\ \hline
\multirow{3}{*}{\bf \#tickets} & train                & 359,012 & 309,635 & 300,358 & 209,580 \\ 
 & validation              & 63,049  & 54,407  & 52,788  & 36,887  \\ 
& test & 50,668 & 9,522 & 22,784 & 15,044 \\ \hline
\multirow{1}{*}{\bf \#intents} &  total         & 683    & 461    & 421    & 281    \\ \hline
\multirow{3}{*}{\bf \#clients} &  train   & 646    & 576    & 639    & 412    \\ 
& validation & 412    & 378    & 401    & 243      \\ 
& test & 394 & 127 & 259 & 4 \\
\end{tabular}%
}
\end{table}

Our approach requires a list of relevant intents for each client as input. Collecting this information for clients is an on-going process and, thus, we utilize 
historical 
predictions of a model 
as a proxy for the client's relevant intents. 
Furthermore, we introduce a coverage parameter to limit the size of the relevant intents lists. For example, if we set the coverage to 100\%, all intents predicted by the production model are part of the list of relevant intents, while reducing it to 99\% ensures that at least 99\% of the tickets are covered by the most frequent intents. This coverage parameter allow us to create different set of relevant intents with different sizes. To achieve this, we sort the intents by frequency and remove tickets with the intent of lowest frequency until we reach the desired coverage. 
Table~\ref{tab:list_intents_details} presents details on the median and maximum number of intents per client based on different coverage levels. Similar to the generic model, the generic model with the list of relevant intents is trained using all training tickets and evaluated on the full generic test set as well as the industry-specific test sets.


\begin{table}[t]
\centering
\caption{Number of relevant intents per client based on different levels of ticket coverage.}
\label{tab:list_intents_details}
\resizebox{\columnwidth}{!}{%
\begin{tabular}{rl|rrrrr}
 \multicolumn{1}{r}{} & &
  \textbf{100\%} &
  \textbf{99\%} &
  \textbf{98\%} &
  \textbf{97\%} &
  \textbf{96\%} \\ \hline
  \multirow{2}{*}{\bf \#intents p/ client} &
 median &
  264 &
  155 &
  123 &
  104 &
  91 \\
  &
 max &
  455 &
  304 &
  255 &
  224 &
  200 \\ 
\end{tabular}%
}
\end{table}

\subsection{Experimental setup}\label{sec:hyperparameters}

All models are XLM-RoBERTa-base models, which contain 125M parameters.
We train all models using cross-entropy loss with a batch size of 512, for a maximum of 30 epochs with early-stopping based on the validation loss and a patience of 3 epochs. We use the AdamW optimizer with an initial learning rate of 1e-6 and a weight decay of 10\%. To reduce training time we use adapters in the last 3 layers of the model, following Pfeiffer's configuration~\cite{pfeiffer2020AdapterHub}. In the classification head we use concatenation as the aggregation function. We set the embedding dropout to 90\%. In the linear projection layer the dimensionality is reduced to 128. Finally, we use just one residual layer.

To ensure robustness, all performances reported ahead are obtained by averaging the individual performance obtained using 4 different training seeds. Furthermore, we perform statistical analysis to determine statistically significant differences between model performances. We use the ranx library \cite{ranx} to extract 1,000 random subsets from the test set. We evaluate the performance of the models on these subsets and compute the p-values using the paired \mbox{t-Test}.
In all comparisons, we assess the performance of each approach against the baseline without using the relevant intents as features. Statistical significance is determined when the \mbox{p-value} is below 0.001.

\subsection{Generic model with relevant intents vs Industry models}\label{sec:generic_vs_industry}

\begin{table}[b!]
\centering
\caption{Generic model versus industry-specific models accuracy on each test set. $^{\dag}$ indicates statistically significant improvements against the corresponding industry-specific model with p-value $\leq$ 0.001.}

\label{tab:generic_vs_industry}
\resizebox{0.49\textwidth}{!}{%
\begin{tabular}{l|rrrr}

 & \multicolumn{4}{c}{\bf Accuracy on the test set} \\ 
\multicolumn{1}{l|}{\multirow{1}{*}{\bf Model}} & \multicolumn{1}{r|}{\bf Generic} & \multicolumn{1}{r|}{\bf Software} & \multicolumn{1}{r|}{\bf E-commerce } & \multicolumn{1}{r}{\bf Finance} \\ \hline
Industry-specific & -- & \multicolumn{1}{|r|}{72.4\%} & \multicolumn{1}{r|}{60.5\%} & \multicolumn{1}{|r}{66.4\%} \\ 
Generic & \multicolumn{1}{|r|}{65.7\%} & \multicolumn{1}{|r|}{72.4\%} & \multicolumn{1}{r|}{60.1\%} & \multicolumn{1}{|r}{66.3\%} \\  
Generic with search & \multicolumn{1}{|r|}{\textbf{67.1\%}} & \multicolumn{1}{|r|}{$^{\dag}$\textbf{76.1\%}} & \multicolumn{1}{r|}{$^{\dag}$\textbf{60.9\%}} & \multicolumn{1}{|r}{\textbf{66.5\%}} \\
\end{tabular}%
}
\end{table}

We start by comparing the performance of the generic model (\mbox{\figref{fig:approaches} \subref{fig:approaches-a}}) against the industry-specific models (\mbox{\figref{fig:approaches} \subref{fig:approaches-c}}). As maintaining multiple industry-specific models incurs high costs, it is desirable to have a single model that can effectively handle tickets from various industries. 
To make this comparison, 
we train three industry-specific models specifically for the Software, E-commerce, and Finance industries. It is important to note that the training data for these models consists of tickets from the entire dataset that are associated with intents belonging to their respective industries. And it should be noted that these subsets may not be entirely disjoint since there are certain intents that may be relevant to multiple industries.

Table~\ref{tab:generic_vs_industry} presents the average accuracy the models across four different training seeds. We observe that the generic model performs on par with the industry-specific models in their respective industries, with only a slight decrease in performance observed for the E-commerce subset. This outcome supports the notion that a single generic model can effectively handle the intent detection task.  Additionally, we introduce the results of the \emph{generic model with search} setting, which incorporates a filtering component (similar to \mbox{\figref{fig:approaches} \subref{fig:approaches-d}}) that selects the most suitable model prediction from the list of relevant intents, as explained in Section~\ref{sec:proposed_solution}. We observe improved performance
when employing this strategy, which passes the significant test for all subset except Finance. For the remainder of this work, we assume that the filtering mechanism consistently outputs ``incorrect" if the model prediction falls outside the list of relevant intents, as described in Section~\ref{sec:proposed_solution}. Furthermore, since we already assessed the performance of the generic model across industries, our subsequent experiments only concern results on the more complete generic set.


\begin{table}[b]
\centering
\caption{Impact of the relevant intents lists' training coverage. $^{\dag}$ indicates statistically significant improvements against the baseline (generic w/filter) with p-value $\leq$ 0.001.}
\label{tab:generic_lists}
\resizebox{0.49\textwidth}{!}{%
\begin{tabular}{lr|rrrrrr|r}
\multicolumn{1}{l}{\multirow{2}{*}{\bf Model}} & {\bf train} & \multicolumn{6}{c}{\bf Accuracy on the test set by test coverage} \\
 &  {\bf coverage} & \multicolumn{1}{c}{\bf  100\%} & \multicolumn{1}{c}{\bf  99\%} & \multicolumn{1}{c}{\bf  98\%} & \multicolumn{1}{c}{\bf  97\%} & \multicolumn{1}{c|}{\bf  96\%} & \multicolumn{1}{c}{\bf 
 Average} \\ \hline
\multicolumn{1}{l}{Generic w/ filter} & -- & \multicolumn{1}{r}{65.7\%} & \multicolumn{1}{r}{64.4\%} & \multicolumn{1}{r}{63.5\%} & \multicolumn{1}{r}{62.1\%} & \multicolumn{1}{r|}{60.9\%} & \multicolumn{1}{|r}{63.3\%} \\ \hline
 & 100\% & \multicolumn{1}{r}{$^{\dag}$3.5pp} & \multicolumn{1}{r}{0.3pp} & \multicolumn{1}{r}{-0.8pp} & \multicolumn{1}{r}{-1.2pp} & -1.5pp & \multicolumn{1}{|r}{+0.1pp} \\
& 99\% & \multicolumn{1}{r}{-0.2pp} & \multicolumn{1}{r}{$^{\dag}$3.0pp} & \multicolumn{1}{r}{$^{\dag}$2.4pp} & \multicolumn{1}{r}{$^{\dag}$1.7pp} & $^{\dag}$1.1pp & \multicolumn{1}{|r}{\textbf{$^{\dag}$+1.6pp}} \\
Generic w/ filter  & 98\% & \multicolumn{1}{r}{-1.9pp} & \multicolumn{1}{r}{$^{\dag}$2.3pp} & \multicolumn{1}{r}{$^{\dag}$2.9pp} & \multicolumn{1}{r}{$^{\dag}$2.5pp} & $^{\dag}$2.2pp & \multicolumn{1}{|r}{\textbf{$^{\dag}$+1.6pp}} \\
and intents & 97\% & \multicolumn{1}{r}{-2.3pp} & \multicolumn{1}{r}{$^{\dag}$1.5pp} & \multicolumn{1}{r}{$^{\dag}$2.5pp} & \multicolumn{1}{r}{$^{\dag}$2.6pp} & $^{\dag}$2.2pp & \multicolumn{1}{|r}{$^{\dag}$+1.3pp} \\
& 96\% & \multicolumn{1}{r}{-3.3pp} & \multicolumn{1}{r}{$^{\dag}$1.2pp} & \multicolumn{1}{r}{$^{\dag}$2.2pp} & \multicolumn{1}{r}{$^{\dag}$2.6pp} & $^{\dag}$2.5pp & \multicolumn{1}{|r}{$^{\dag}$+1.0pp} \\ 
\end{tabular}%
}
\end{table}

\subsection{Robustness}\label{sec:robustness}


\begin{table}[b]
\centering
\caption{Impact of training with noise in the relevant intents lists. The coverage of the intents list in training is fixed at 98\%. $^{\dag}$ indicates statistically significant improvements against the baseline (generic w/filter) with p-value $\leq$ 0.001.}
\label{tab:models_noise}
\resizebox{0.49\textwidth}{!}{%
\begin{tabular}{lr|rrrrrrr}
\multicolumn{1}{l}{\multirow{2}{*}{\bf Model}} & {\bf train} & \multicolumn{6}{c}{\bf Accuracy on the test set by test coverage} \\
 &  {\bf noise} & \multicolumn{1}{c}{\bf  100\%} & \multicolumn{1}{c}{\bf  99\%} & \multicolumn{1}{c}{\bf  98\%} & \multicolumn{1}{c}{\bf  97\%} & \multicolumn{1}{c|}{\bf  96\%} & \multicolumn{1}{c}{\bf 
 Average} \\ \hline
\multicolumn{1}{l}{Generic w/ filter} & -- & \multicolumn{1}{r}{65.7\%} & \multicolumn{1}{r}{64.4\%} & \multicolumn{1}{r}{63.5\%} & \multicolumn{1}{r}{62.1\%} & \multicolumn{1}{r|}{60.9\%} & \multicolumn{1}{|r}{63.3\%} \\ \hline
 & 0\% & \multicolumn{1}{r}{-1.9pp} & \multicolumn{1}{r}{$^{\dag}$2.3pp} & \multicolumn{1}{r}{$^{\dag}$2.9pp} & \multicolumn{1}{r}{$^{\dag}$2.5pp} & \multicolumn{1}{r|}{$^{\dag}$2.2pp} & $^{\dag}$+1.6pp \\
 & 5\% & \multicolumn{1}{r}{$^{\dag}$0.6pp} & \multicolumn{1}{r}{$^{\dag}$2.2pp} & \multicolumn{1}{r}{$^{\dag}$2.4pp} & \multicolumn{1}{r}{$^{\dag}$2.1pp} & \multicolumn{1}{r|}{$^{\dag}$1.7pp} & \textbf{$^{\dag}$+1.8pp} \\
Generic w/ filter & 10\% & \multicolumn{1}{r}{$^{\dag}$0.4pp} & \multicolumn{1}{r}{$^{\dag}$1.5pp} & \multicolumn{1}{r}{$^{\dag}$1.6pp} & \multicolumn{1}{r}{$^{\dag}$1.4pp} & \multicolumn{1}{r|}{$^{\dag}$0.9pp} & $^{\dag}$+1.2pp \\
 and intents & 20\% & \multicolumn{1}{r}{$^{\dag}$0.3pp} & \multicolumn{1}{r}{$^{\dag}$0.7pp} & \multicolumn{1}{r}{$^{\dag}$0.8pp} & \multicolumn{1}{r}{$^{\dag}$0.6pp} & \multicolumn{1}{r|}{$^{\dag}$0.4pp} & $^{\dag}$+0.6pp \\
& 50\% & \multicolumn{1}{r}{-0.3pp} & \multicolumn{1}{r}{-0.2pp} & \multicolumn{1}{r}{-0.2pp} & \multicolumn{1}{r}{-0.2pp} & \multicolumn{1}{r|}{-0.2pp} & -0.2pp \\ 
\end{tabular}%
}
\end{table}

The list of relevant intents for a client provides valuable information to improve model performance. However, relevant intents
can evolve over time, underscoring the importance of ensuring the production model's robustness to such changes without the need of model retraining. Retraining the model every time the list of client relevant intents wants to be modified incurs significant costs. Thus, it becomes imperative to develop a production model that can effectively handle variations in the list of relevant intents without requiring frequent re-deployment. This not only saves resources but also enhances the model's overall efficiency and scalability.

To assess the impact of changes in the list of relevant intents on performance, we train generic models with relevant intents lists using different coverage values, such as 100\%, 99\%, etc. Then, we evaluate these models by providing input intents lists with varying coverages. This experimental setup simulates scenarios where clients add or remove relevant intents over time. 
%
In the first row of Table~\ref{tab:generic_lists}, we show the performance of the baseline generic model 
with the output filter, that ensures outputs within relevant intents. 
Note that, although it does not utilize the list of
intents as input, the performance of the baseline also decreases
due to the filtering mechanism, which classifies predictions outside of the list of relevant intents as incorrect. The subsequent rows show the delta in accuracy when comparing the generic model trained with different coverages of relevant intents lists against the performance of baseline generic model on the same coverage level. 


Overall, our results indicate that incorporating the list of relevant intents as features leads to performance improvements, as evidenced by the average gains across various coverages when compared to the generic model. We also observe that the gains in performance are statistically significant for most direct comparisons. However, it is noteworthy that performance tends to deteriorate when the training coverage differs from the testing coverage.
For instance, training and evaluating the model with a coverage of 100\% achieves an accuracy of 69.2\%. However, training with the same coverage and evaluating with a list coverage of 96\% results, the accuracy is 59.4\%, which is a drop of $\approx10\%$. 
While some of the expected degradation was reduced through the use of high values of dropout ($90\%$) in the embedding layer of the relevant intents, 
these results can still raise some concerns regarding potential overfitting.

In an attempt to enhance the model's resilience to changes in the relevant intents lists, we introduce synthetic noise during the training process. The hypothesis is that by exposing the model to different intents lists for the same client during training it will become more resistant to real-world changes in the lists. This means that the same client may have different relevant intents lists during the training phase. To introduce the noise in the relevant intents lists, we change $k$\% of the intents present in the list, e.g, if an intent is not in the relevant intents list for an client, we add it with probability of $k$\%, and the reverse is done if the intent is present in the list. We train models with the following noise values: 0\% (no noise), 5\%, 10\%, 20\% and 50\%. To reduce the number of combinations, we fix the training coverage to 98\%, we then evaluate those models on different coverage values to assess the models' robustness regarding list changes. We chose 98\% coverage since it is the setting with highest average performance (from Table~\ref{tab:generic_lists}) and because we can analyse its performance when removing intents (e.g., 97\% and 96\% coverages) and also when adding intent (e.g., 99\% and 100\% coverages).

From the results in Table~\ref{tab:models_noise} we observe that the model trained without any training noise in the relevant intents lists shows high degradation in terms of accuracy when subject to the changes in the coverage values: when the coverage is 98\%, the model's accuracy is 66.4\%, which drops to 63.8\% when the coverage is 100\%. This indicates that simply training a model where the lists are the ones seen during training is not robust to subsequent changes in the relevant intents lists.
The models trained with 5\% noise are the most robust to changes in the lists and have high performance overall. Models trained with 10\% and 20\% also show gains in robustness but notable drops in accuracy, while the model trained with 50\% noise in the relevant intents list has very similar (but lower) performance to the generic model without relevant intents lists, suggesting that the model simply learned to ignore the relevant intents lists and rely only on the ticket content. 
We show the delta in performance of our approach versus the baseline generic model and highlight statistical significant results. 

\begin{table}[b!]
\centering
\caption{Out-of-domain evaluation results comparing the generic model with or without relevant intents and with or without intents training noise. 
$^{\dag}$ indicates statistically significant improvements against the baseline (generic w/ filter) with p-value $\leq$ 0.001.}
\label{tab:ood_evaluation}
\resizebox{0.95\columnwidth}{!}{%
\begin{tabular}{l|rrr}
 & \multicolumn{2}{c}{\bf Accuracy on the test set}              \\
{\bf Model}   & \multicolumn{1}{r|}{\bf in-domain}    & {\bf out-of-domain} \\
\hline
Generic w/ filter                 & \multicolumn{1}{r|}{63.3\%}               & 45.0\% \\ 
Generic w/ filter and intents, no noise  & \multicolumn{1}{r|}{$^{\dag}$64.9\%} & $^{\dag}$45.4\% \\ 
Generic w/ filter and  intents, 5\% noise & \multicolumn{1}{r|}{\bf $^{\dag}$65.1\%}  & {\bf$^{\dag}$45.9\%} \\ 
\end{tabular}%
}
\end{table}

\subsection{Out-of-domain evaluation}\label{sec:odd}

As a final experiment, we further assess our models’ robustness by evaluating them on tickets from out-of-domain
clients; i.e., from clients whose tickets were not seen during training.
To simulate that, we create an out-of-domain split by segregating all tickets from a few clients as test, and leaving the other clients for train and validation. the final distribution of tickets in the out-of-domain dataset is of 77.7\%, 11.7\% and 10.6\% for the train, validation and test splits, respectively. 
 

We train and evaluate 
models in the out-of-domain context, namely one generic model without any extra information and two generic models with relevant intents lists, one trained without noise in the list of relevant intents and the other one trained with 5\% noise in the intents. The list of relevant intents is fixed to have 98\% training coverage and the evaluation performed on lists with 100\%, 99\%, 98\%, 97\%, and 96\% coverage, and we average out the results across all coverages, like we did for the 
previous
experiment. 
We also include the previous results on the in-domain sets for comparison.

We observe that the models trained with the list of relevant intents perform better than the generic model across both dataset splits (Table~\ref{tab:ood_evaluation}). This result indicates that the list of relevant intents helps actually boost the model performance by keeping its ability to generalize and, thus, the model is not simply learning to memorize the input lists of relevant intents. We also notice that adding training noise further increases performance, highlighting the gains in model robustness.

\section{Related work}
\label{sec:related_work}


In the customer support scenario, machine learning methods have been widely explored to improve the analysis and handling of requests. Some examples include, using deep learning models to assign issues to the appropriate workers~\cite{feng2022tadaa}, exploring ensembles of deep learning architectures to classify customer support tickets~\cite{zicari2021discovering}, or  
classifying incoming emails using machine learning models and directing them to predefined queues based on their topic~\cite{borg2021mail}, allowing agents to select queues that align with their field of expertise. Here we are only concerned with the task of automatically detecting intents and not of matching tickets to predefined queues, nor with aligning agents with priority queues. 
Furthermore, none of cited works explore the practical consideration of scaling a system in production for various clients, considering a solution that accounts for costs, maintenance, usability expectations and performance quality.

Transformer-based architectures~\cite{vaswani2017attention} like BERT~\cite{devlin2018bert}
have achieved state-of-the-art results in NLP classification tasks, making them suitable for intent detection. 
XLM-RoBERTa~\cite{conneau2019unsupervised} is particularly relevant as tickets can be in languages other than English. Other architectures were also explored for intent detection including
dual-encoders~\cite{casanueva2020efficient} and recurrent neural networks (RNNs)\cite{guo2014joint}.
In particular, Casanueva et al. \cite{casanueva2020efficient} employed a setup which used the combination of two dual encoders
to efficiently train a multi-layer perceptron on top of the representations given by the encoder models. Chen et al.\cite{chen2019bert} utilized the BERT model\cite{devlin2018bert} to perform joint intent classification and slot filling, using the output embedding of the special \texttt{[CLS]} token for intent classification.

In scenarios where inputs consist of different types of data, multi-modal machine learning methods have gained popularity. These approaches handle inputs that combine text, numerical features, audio, and other modalities
~\cite{ngiam2011multimodal}.
Here, we are mainly concerned with merging textual inputs with a list of integers representing the relevant intents. In the context of customer support, tickets often contain valuable non-textual information alongside the text message, which can enhance natural language understanding tasks. A common approach is to use a Transformer encoder, such as BERT~\cite{devlin2018bert}, to process the text features and extract the embedding of the special \texttt{[CLS]} token, which captures information from the entire text sequence. This representation is then combined with vectorized non-text features and fed into a feed-forward network, we employ a similar approach.

\section{Conclusions}
\label{sec:conclusions}
We introduce a new system to perform intent classification of customer requests
that accurately 
scales with the number of clients and business. 
We developed an architecture that utilizes a generic model augmented with a per-client relevant intents list. Our approach eliminates the need for assigning clients to specific industries and reduces maintenance costs by deploying a single model. Furthermore, the incorporation of relevant intents lists obtained directly from clients allows for a more personalized experience that can easily adapt to changes in clients' needs.

In terms of performance, our approach surpasses industry-specific and generic models. 
We further improve our model's performance by augmenting it with a list of relevant intents as features and show that the list of relevant intents is robust to changes when trained with added synthetic noise. 
Finally, we conducted an out-of-domain evaluation using a testing set that included tickets from clients unseen during training and conclude that our model effectively utilizes the relevant intents lists as valuable information for classification, rather than relying solely on them as client identifiers.



\bibliographystyle{abbrv}
\bibliography{main}

\begin{thebibliography}{10}

\bibitem{ranx}
E.~Bassani.
\newblock ranx: {A} blazing-fast python library for ranking evaluation and
  comparison.
\newblock In {\em {ECIR} {(2)}}, volume 13186 of {\em Lecture Notes in Computer
  Science}, pages 259--264. Springer, 2022.

\bibitem{borg2021mail}
A.~Borg, M.~Boldt, O.~Rosander, and J.~Ahlstrand.
\newblock E-mail classification with machine learning and word embeddings for
  improved customer support.
\newblock {\em Neural Computing and Applications}, 33(6):1881--1902, 2021.

\bibitem{casanueva2020efficient}
I.~Casanueva, T.~Tem{\v{c}}inas, D.~Gerz, M.~Henderson, and I.~Vuli{\'c}.
\newblock Efficient intent detection with dual sentence encoders.
\newblock In {\em Proceedings of the 2nd Workshop on Natural Language
  Processing for Conversational AI}, pages 38--45. Association for
  Computational Linguistics, July 2020.

\bibitem{chen2019bert}
Q.~Chen, Z.~Zhuo, and W.~Wang.
\newblock Bert for joint intent classification and slot filling.
\newblock {\em arXiv preprint arXiv:1902.10909}, 2019.

\bibitem{conneau2019unsupervised}
A.~Conneau, K.~Khandelwal, N.~Goyal, V.~Chaudhary, G.~Wenzek, F.~Guzm{\'a}n,
  E.~Grave, M.~Ott, L.~Zettlemoyer, and V.~Stoyanov.
\newblock Unsupervised cross-lingual representation learning at scale.
\newblock In {\em Proceedings of the 58th Annual Meeting of the Association for
  Computational Linguistics}, pages 8440--8451, Online, July 2020. Association
  for Computational Linguistics.

\bibitem{devlin2018bert}
J.~Devlin, M.-W. Chang, K.~Lee, and K.~Toutanova.
\newblock {BERT}: Pre-training of deep bidirectional transformers for language
  understanding.
\newblock In {\em Proceedings of the 2019 Conference of the North {A}merican
  Chapter of the Association for Computational Linguistics: Human Language
  Technologies, Volume 1 (Long and Short Papers)}, pages 4171--4186, 2019.

\bibitem{feng2022tadaa}
L.~Feng, J.~Senapati, and B.~Liu.
\newblock Tadaa: real time ticket assignment deep learning auto advisor for
  customer support, help desk, and issue ticketing systems.
\newblock {\em arXiv preprint arXiv:2207.11187}.

\bibitem{guo2014joint}
D.~Guo, G.~Tur, W.-t. Yih, and G.~Zweig.
\newblock Joint semantic utterance classification and slot filling with
  recursive neural networks.
\newblock In {\em 2014 IEEE Spoken Language Technology Workshop (SLT)}, pages
  554--559. IEEE, 2014.

\bibitem{liu2019roberta}
Y.~Liu, M.~Ott, N.~Goyal, J.~Du, M.~Joshi, D.~Chen, O.~Levy, M.~Lewis,
  L.~Zettlemoyer, and V.~Stoyanov.
\newblock Roberta: A robustly optimized bert pretraining approach.
\newblock {\em arXiv preprint arXiv:1907.11692}.

\bibitem{mesquita2022dense}
T.~Mesquita, B.~Martins, and M.~Almeida.
\newblock Dense template retrieval for customer support.
\newblock In {\em Proceedings of the 29th International Conference on
  Computational Linguistics}, pages 1106--1115, 2022.

\bibitem{ngiam2011multimodal}
J.~Ngiam, A.~Khosla, M.~Kim, J.~Nam, H.~Lee, and A.~Y. Ng.
\newblock Multimodal deep learning.
\newblock In {\em Proceedings of the 28th international conference on machine
  learning (ICML-11)}, pages 689--696, 2011.

\bibitem{pfeiffer2020AdapterHub}
J.~Pfeiffer, A.~R{\"u}ckl{\'e}, C.~Poth, A.~Kamath, I.~Vuli{\'c}, S.~Ruder,
  K.~Cho, and I.~Gurevych.
\newblock Adapterhub: A framework for adapting transformers.
\newblock In {\em Proceedings of the 2020 Conference on Empirical Methods in
  Natural Language Processing: System Demonstrations}, pages 46--54, 2020.

\bibitem{vaswani2017attention}
A.~Vaswani, N.~Shazeer, N.~Parmar, J.~Uszkoreit, L.~Jones, A.~N. Gomez,
  {\L}.~Kaiser, and I.~Polosukhin.
\newblock Attention is all you need.
\newblock {\em Advances in neural information processing systems}, 30, 2017.

\bibitem{zicari2021discovering}
P.~Zicari, G.~Folino, M.~Guarascio, and L.~Pontieri.
\newblock Discovering accurate deep learning based predictive models for
  automatic customer support ticket classification.
\newblock In {\em Proceedings of the 36th annual ACM symposium on applied
  computing}, pages 1098--1101, 2021.

\end{thebibliography}

\end{document}